\documentclass{article}

\PassOptionsToPackage{numbers, compress}{natbib}
\usepackage[preprint]{neurips_2026}

\usepackage[utf8]{inputenc} 
\usepackage[T1]{fontenc}    
\usepackage{hyperref}       
\usepackage{url}            
\usepackage{booktabs}       
\usepackage{amsfonts}       
\usepackage{nicefrac}       
\usepackage{microtype}      
\usepackage{xcolor}         
\usepackage{amsmath}
\usepackage{amssymb} 
\usepackage{booktabs}
\usepackage{colortbl} 
\usepackage{array}
\usepackage{wrapfig}
\usepackage{graphicx}
\usepackage{multirow}

\usepackage{tcolorbox}
\tcbuselibrary{listings,skins,breakable}  

\title{Alignment Is All You Need: Instruction-Free Training for General Audio-Language Models}

\author{
  Xuanru Zhou$^{1,2}$\thanks{Work done during an internship at Tencent Hunyuan.} \quad Yiwen Shao$^{2}$\thanks{Corresponding author.} \quad Jiahong Li$^{2}$ \quad Dong Yu$^{2}$ \\[0.4em]
  $^{1}$Zhejiang University \quad $^{2}$Tencent Hunyuan \\[0.3em]
  \texttt{xuanruzhou15@gmail.com} \ \ \ \ \   \texttt{yshao18@jhu.edu} \\[0.5em]
  \textbf{Code}: \url{https://github.com/rorizzz/IFAO-lalm} \\[0.3em]
  \textbf{Hugging Face}: \url{https://huggingface.co/collections/eureka1500/ifao-lalm}
  }

\begin{document}

\maketitle

\begin{abstract}
Multimodal large language models (MLLMs) are typically built through a multi-stage pipeline consisting of cross-modal alignment, supervised fine-tuning (SFT), and preference optimization.
This pipeline assumes that adapting an LLM to a new modality requires extensive task-specific supervision.
However, pretrained LLMs already possess strong reasoning and instruction-following abilities. 
As LLMs evolve rapidly, an important question remains: can we efficiently transfer these capabilities to a new modality with minimal intervention, and is alignment alone sufficient for building a multimodal model?
We introduce an \textbf{Instruction-Free Alignment-Only} large audio-language model (LALM) that keeps both the audio encoder and the LLM fully frozen, learning only a lightweight projector. 
Borrowing insights from AZeroS~\cite{shao2026azeros}, we train on \textit{(audio, response)} pairs from Self-Generated Data Construction, where an LLM expands captions into free-form responses without explicit task instructions.
Across MMAU, MMAR, MMSU, and MMAU-Pro, our approach matches or surpasses heavily post-trained baselines using substantially less data.
By keeping the LLM frozen, our model preserves its native instruction-following competence and can port seamlessly across model generations.
Our results suggest that competitive MLLM can emerge from alignment alone, reducing multimodal extension to a lightweight projector-training problem that generalizes across modalities and adapts rapidly to each new LLM release.

\end{abstract}

\section{Introduction}

\begin{figure*}[t]
\label{fig:main}
    \centering
\resizebox{\textwidth}{!}{\includegraphics{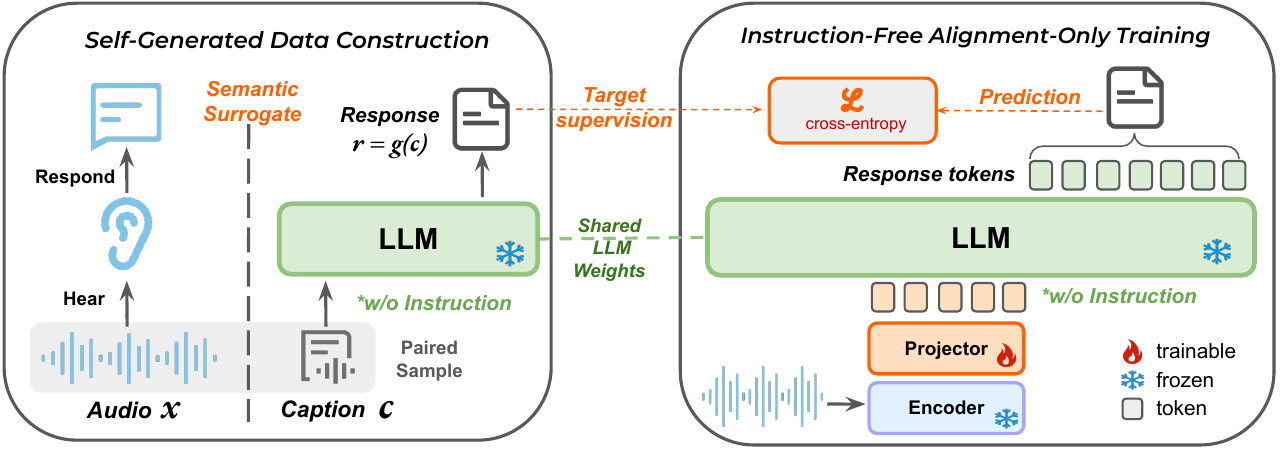}}
    \caption{\textbf{Overview of the pipeline.} \textit{Left: Self-Generated Data Construction.} The dashed line separates two views: on the left, the real listening process, where a human hears audio $x$ and responds; on the right, our generation surrogate. Instead of collecting human responses, we feed the paired caption $c$ into the frozen LLM without any instruction to obtain $r = g(c)$. The caption thus serves as a \textit{semantic surrogate} for the audio, and $r$ becomes the training target. 
    \textit{Right: Instruction-Free Alignment-Only training.} Audio $x$ passes through a frozen encoder and a trainable projector into the same frozen LLM, again without instructions. The LLM here is identical to the one used on the left. This consistency ensures that $r$ matches what this LLM would produce given the caption surrogate. Training the projector with cross-entropy against $r$ therefore aligns audio representations to the LLM's own caption-conditioned response distribution, not to external annotation.}
    \vspace{-5mm}
\end{figure*}

Large language models (LLMs) pretrained on massive text corpora already exhibit strong reasoning, instruction-following, and generalization abilities~\citep{browngpt3, openai2024gpt4technicalreport, grattafiori2024llama3herdmodels, qwen2025qwen25technicalreport, wei2022emergent}.
Instruction tuning further turns these models into a \emph{universal decoder}: a single model capable of following arbitrary text instructions in a zero-shot manner~\citep{ouyang2022training, wei2022finetuned, sanh2022multitask, chung2024scaling}. 
Modern multimodal large language models (MLLMs) extend this capability to non-text modalities using a common architecture: a pretrained modality encoder, a learnable projector, and a pretrained LLM~\citep{alayrac2022flamingo, li2023blip2bootstrappinglanguageimagepretraining, liu2023llava}. 
Training typically consists of three stages: (1) cross-modal alignment, (2) supervised fine-tuning (SFT), and (3) preference optimization~\citep{ouyang2022training, christiano2017deep, rafailov2023dpo}. 
Large audio-language models (LALMs) follow the same recipe, from early systems~\citep{gong2024listen, deshmukh2023pengi} to recent large-scale models spanning speech, music, and environmental sound~\citep{tang2024salmonn, xu2025qwen25omnitechnicalreport, af2, af3, kimiteam2025kimiaudiotechnicalreport}. 

In current practice, most performance gains are attributed to SFT and preference optimization, while cross-modal alignment is treated mainly as an initialization stage. 
However, these later stages improve task performance by adapting the LLM toward the post-training instruction distribution. As the model becomes increasingly specialized to task-specific supervision, its original universal decoding behavior can gradually erode~\citep{Kirkpatrick_2017, luo2023empirical}.
Moreover, every new LLM generation requires rerunning expensive post-training pipelines on a fresh backbone. 
This raises a simple question: \emph{is alignment alone sufficient?} 
If the pretrained encoder already captures audio information, and the pretrained LLM already knows how to follow instructions, then perhaps multimodal training can be reduced to efficiently connecting the two, rather than retraining either side.

In this work, we show that alignment alone is indeed sufficient for building \textit{competitive} LALMs. 
Our approach keeps both the audio encoder and the LLM fully frozen, and trains only a lightweight projector. 
Following the \textbf{Instruction-Free} paradigm of~\citet{shao2026azeros}, we remove task instructions entirely during training, forcing the projector to align audio representations directly to the LLM's latent semantic space. 
To support this training setup, we build upon \textbf{Self-Generated Data Construction} framework of~\citet{shao2026azeros}, extending it from speech to general audio.
Instead of manually designing QA pairs or instruction templates, we treat captions as \textit{semantic surrogates} for audio and use an LLM to automatically expand them into free-form responses. 
The resulting pipeline is fully automatic, requiring no human-written instructions, task taxonomy, or manually curated supervision.

Our study leads to three main findings. 
First, instruction-free alignment-only training generalizes surprisingly well, matching or surpassing heavily post-trained LALMs on MMAU~\cite{MMAU}, MMAR~\cite{ma2025mmar}, MMSU~\cite{wang2025mmsu}, and MMAU-Pro~\cite{kumar2025mmau-pro} using substantially less data. 
Second, because the LLM itself remains frozen, its pretrained instruction-following capability is preserved, leading to state-of-the-art performance on the instruction-following split of MMAU-Pro among open-source LALMs. 
Third, reachable performance is jointly bounded by encoder informativeness and LLM capability. Scaling alignment data mainly benefits tasks requiring richer open-ended reasoning, while closed-set recognition quickly saturates once encoder limitations dominate. 
The saturation point itself depends on data coverage and diversity, which determine how fully the alignment approaches this bound. Sweeping encoders, the best-performing model on each benchmark consistently tracks the capability emphasized during encoder pretraining, while sweeping LLMs shows that the recipe transfers cleanly across model generations under a matched-generator condition.
Finally, we establish that our approach does not strictly depend on dense synthetic captions per sample; rather, simple open-source captions are sufficient when aggregated across diverse datasets.

\vspace{-3mm}
\paragraph{Contributions.}
First, we present an \textbf{Instruction-Free Alignment-Only} LALM that matches or surpasses heavily post-trained baselines on MMAU, MMAR, MMSU, and MMAU-Pro using substantially less data.
Second, we demonstrate that competitive multimodal instruction following can emerge from frozen encoders and frozen LLMs, without SFT or preference optimization.
Third, we extend the Self-Generated Data Construction framework of~\citet{shao2026azeros} from speech to general audio, enabling a fully automatic training pipeline without manually curated QA pairs, task templates, or instruction taxonomies.
Fourth, we provide extensive encoder and LLM ablations showing that performance is jointly constrained by encoder representations and LLM capability, rather than data scaling alone.
We have released the code, datasets, and model weights to support future research on alignment-only audio-language modeling.

\section{Related Work}
\label{sec:related-work}

\subsection{Multimodal Alignment and Instruction Tuning}
The development of Multimodal Large Language Models (MLLMs) has established a robust two-stage paradigm: initial modality alignment followed by Task-Specific Instruction Tuning (TSIT)~\citep{li2023blip2bootstrappinglanguageimagepretraining, liu2023llava}. 
Recent methods align model responses via human intent through Reinforcement Learning from Human Feedback (RLHF)~\citep{NEURIPS2022_b1efde53}, with Direct Preference Optimization (DPO)~\citep{rafailov2023dpo} offering a more direct alternative.
This paradigm has been extensively adopted in the audio domain to build Large Audio-Language Models (LALMs). Early efforts like Qwen-Audio~\cite{Qwen-Audio} and Qwen2-Audio~\cite{Qwen2-Audio} utilize hierarchical tags or natural-language prompts to unify diverse tasks such as ASR, sound event detection, and music analysis. Similarly, SALMONN~\cite{tang2024salmonn} and WavLLM~\cite{hu-etal-2024-wavllm} employ dual-encoder architectures and massive TSIT datasets to achieve competitive performance across predefined audio tasks.
However, these models often exhibit a form of \textit{instruction-induced bias}, where they overfit to the distribution of training prompts and task formats, leading to responses that reflect specific templates rather than the underlying input signal~\cite{wei2022finetuned, sanh2022multitask, min-etal-2022-rethinking}.
In the audio domain, this bias is often reflected in a tendency to default to ASR-style transcriptions, even for non-speech inputs~\cite{zhang-etal-2023-speechgpt, audiogpt}.

To overcome the limitations of TSIT, recent research shifts toward leveraging the inherent generalization ability of pretrained LLMs, rather than relying on explicit task-specific supervision.
AudioChatLLaMA~\cite{fathullah-etal-2024-audiochatllama} aligns speech and text by enforcing consistent LLM responses across modalities, while BLSP \citep{wang2024blsp} achieves similar alignment through a continuation-based objective.
To incorporate richer acoustic information, DeSTA2~\cite{desta2} and DeSTA2.5~\cite{lu2026desta25audiogeneralpurposelargeaudio} introduce description-based training, where a frozen LLM generates synthetic targets by combining transcripts with paralinguistic metadata.
Building on this direction, \citet{shao2026azeros} formalizes an instruction-free training paradigm in the \textit{speech} domain to mitigate template-dependency.

\subsection{Audio Representation Learning}

Audio representation learning seeks to build general-purpose models capable of supporting diverse audio understanding tasks, and can be broadly categorized into three paradigms.
Task-specific supervised learning has produced strong encoders for audio event classification~\citep{VGGish, OpenL3, Panns, AST, HTSAT, Dasheng}, speech recognition~\citep{whisper}, and speaker recognition~\citep{xvector, ecapa}, but is inherently limited by predefined label spaces.
Self-supervised learning (SSL) improves generalization by leveraging large-scale unlabeled data~\citep{wav2vec2, hsu2021hubert, wavlm, data2vec, ssast, audioMAE, BEATs, li2022atst, MERT, muq}, though most models remain specialized to particular domains.
Audio-Language Pre-training (ALP) learns richer semantic features by aligning audio with text, typically via contrastive~\cite{CLAP2022, CLAP2023, laionclap2023, guzhov2021audioclip} or generative objectives (e.g., captioning~\citep{tschannen2023cappa,midashenglm7b, af3}). 
However, existing approaches are often constrained by limited task coverage and weak caption supervision. 
Recent work addresses these issues by expanding to full-spectrum audio and improving caption quality, e.g., CaptionStew~\citep{tseng2026revisiting}, which spans speech, music, and environmental sounds, and captioner-generated annotations that provide detailed natural language descriptions~\citep{zhou2026uts}.
In this work, we evaluate our instruction-free alignment-only training across five frozen encoders from these paradigms. This ensures that our findings are robust to the diverse inductive biases inherent in different representation spaces.

\section{Methodology}
\label{sec:methodology}

\subsection{Motivation and Problem Setup}
\label{sec:framework}

A pretrained LLM already follows arbitrary text instructions. Adapting it to a new modality is therefore not a matter of teaching new tasks; it is a matter of presenting modality-specific input the LLM can already interpret~\citep{li2023blip2bootstrappinglanguageimagepretraining, driess2023palme}. Standard MLLM pipelines pursue this goal indirectly, training the full system end-to-end on \textit{(multimodal signal, instruction, response)} triples. Optimizing for specific instructions shapes the encoder toward task-relevant features and updates the LLM's parameters, eroding the \textit{universal-decoder} behavior~\cite{wei2022emergent} it started with. The resulting system's reach is bounded by the instruction corpus, not by the components themselves. 
In this paper, we keep both encoder and LLM frozen and train only a projector that maps audio into the LLM's input embedding space. 
The encoder's audio representations and the LLM's instruction-following are already learned; the projector's job is to connect them, not to retrain either side.

We construct our LALM from three components: a frozen audio encoder $E$, a frozen autoregressive LLM $\mathcal{L}$, and a learnable projector $P_\theta$. The encoder maps a raw audio sample $x \in \mathcal{X}_{\mathrm{audio}}$ to a sequence of feature vectors $E(x) \in \mathbb{R}^{T \times d_E}$, with $T$ scaling with audio duration. With input embedding dimension $d_\mathcal{L}$, the LLM defines a next-token distribution $p_\mathcal{L}(y_t \mid y_{<t}, h)$ given any continuous prefix $h$ in its input embedding space. The projector $P_\theta \colon \mathbb{R}^{d_E} \to \mathbb{R}^{d_\mathcal{L}}$ applies position-wise to the encoder output, and its parameters $\theta$ are the only trainable weights in the system.

Given an audio sample $x$ and target response $y = (y_1, \dots, y_n)$, the projector produces an audio prefix
\begin{equation}
  h \;=\; P_\theta\bigl(E(x)\bigr) \;\in\; \mathbb{R}^{T \times d_\mathcal{L}},
  \label{eq:audio-prefix}
\end{equation}
which the LLM continues autoregressively to generate $y$, with no task instruction prepended.

Standard MLLM training combines three stages: cross-modal alignment, supervised fine-tuning (SFT), and preference optimization. Most reported gains come from stages two and three (Sec. \ref{sec:related-work}). We retain only stage one, on a hypothesis the next two subsections formalize. A well-aligned projector elicits the LLM's pretrained capabilities without further fine-tuning. The system's reach is then jointly bounded by the encoder and LLM, not by any instruction corpus.

\subsection{Self-Generated Data Construction}
\label{sec:data-construction}

To avoid instruction-induced bias, it is essential to establish a target that provides the definitive interpretation of audio.
\citet{shao2026azeros} cast this target as a textual representation for speech, focusing on transcription and paralinguistic attributes.
we expand this scope to the full auditory spectrum, where a \textbf{caption} acts as a \textit{semantic surrogate} that exhaustively describes the audio's content.
From the perspective of a frozen LLM, the semantic content of a prefix - whether provided as text (caption $c$) or as a continuous embedding ($P_\theta(E(x))$ from Eq.~\ref{eq:audio-prefix}) - is functionally equivalent. 
By treating the caption as a semantic surrogate for the audio, we can generate high-quality training targets without manual task templates or human-curated instructions. 

Our construction pipeline consists of two stages. First, a caption source $\mathcal{S}$ provides a natural-language description $c \sim p_\mathcal{S}(\cdot \mid x)$ for each audio $x \in \mathcal{X}$.
We explore two distinct paradigms for $\mathcal{S}$: 
(i) leveraging ground-truth paired text from open-source datasets, and (ii) generating synthetic captions via \textit{Qwen3-Omni-Captioner}~\citep{Qwen3-Omni}.
These sources offer different trade-offs in descriptive granularity and coverage, which we ablate in Sec.~\ref{subsubec:caption}.
Second, the LLM expander $g$ treats the caption $c$ as a textual proxy for the audio signal, generating a fluent response $r = g(c)$ as if it were directly perceiving the audio content. Composing the two stages yields the alignment dataset:
\begin{equation}
  \mathcal{D}_{\mathrm{align}} \;=\; \bigl\{\, (x,\, g(c)) \,:\, x \in \mathcal{X},\ c \sim p_\mathcal{S}(\cdot \mid x) \,\bigr\}.
  \label{eq:d-align}
\end{equation}
This construction is entirely autonomous, requiring no manual task selection or template design.
This approach stands in sharp contrast to standard SFT, which typically organizes data around a finite set of explicit tasks, denoted as $\mathcal{D}_{\mathrm{SFT}} = \{(x, q_i, a_i) : i \in \mathcal{T}\}$. 
In such setups, $\mathcal{T}$ covers discrete categories like sound-event recognition or speaker identification, and model generalization is inherently bounded by the scope of $\mathcal{T}$ encountered during training.
In contrast, $\mathcal{D}_{\mathrm{align}}$ eschews fixed task indices, offering free-form, task-agnostic responses that capture the full semantic essence of the audio.

\subsection{Instruction-Free Alignment-Only Training}
\label{sec:LALM}
The training setup follows directly from Sec. \ref{sec:framework} and Sec. \ref{sec:data-construction}. As input, the LLM consumes only the audio prefix $h = P_\theta(E(x))$ from eq.~\ref{eq:audio-prefix}: no system prompt, no task instruction, no preamble. Supervision is the response $y = (y_1, \ldots, y_n)$ drawn from $\mathcal{D}_{\mathrm{align}}$. 
To ensure distributional consistency, the LLM instance $\mathcal{L}$ used during this alignment phase must be identical to the one employed as the expander $g$ in Sec.~\ref{sec:data-construction}.
We minimize the causal language-modeling loss on response tokens, with gradients flowing only to the projector parameters $\theta$ while $E$ and $\mathcal{L}$ stay frozen:
\begin{equation}
  \mathcal{L}_{\mathrm{align}}(\theta) \;=\; -\,\mathbb{E}_{(x, y) \sim \mathcal{D}_{\mathrm{align}}}\, \sum_{t=1}^{n} \log p_\mathcal{L}\bigl(y_t \mid y_{<t},\, P_\theta(E(x))\bigr).
  \label{eq:align-loss}
\end{equation}
This realizes the \textbf{instruction-free training} regime articulated by \citet{shao2026azeros}. 
By omitting instructions at the input, the audio prefix alone must trigger the desired response, forcing the optimization pressure entirely onto the projector.
Freezing $E$ and $\mathcal{L}$ safeguards their extensive pre-trained priors, particularly preventing the ``erosion'' of the LLM's universal decoding capacity, which is a necessity for maintaining broad generalization without collapsing toward the specific alignment distribution. 
This strategic choice enables an exceptionally \textit{lightweight training paradigm}: by optimizing only the projector, we achieve rapid convergence and competitive performance.

\subsection{Theoretical Analysis}
\label{sec:theoretical-analysis}

The effectiveness of our Instruction-Free Alignment-Only approach rests on the hypothesis that modern LLMs are universal decoders~\cite{wei2022emergent}: they possess the inherent capability to execute nearly any task if provided with a faithful textual representation of the input. Under this view, the goal of multimodal adaptation is not to teach the LLM new reasoning paths, but rather achieving a global \textit{input alignment} that allows the LLM to perceive audio as clearly as it perceives text.

\subsubsection{Alignment Completeness}
\label{sec:theory-alignment}

We formalize the alignment between an audio prefix $h_\theta(x)$ and its textual counterpart $\tilde{h}(c)$ through the lens of response distributions. Two prefixes are \textit{$i$-aligned} if they elicit identical outputs under instruction $i$:
\begin{equation}
h_\theta(x) \sim_i \tilde{h}(c) \Longleftrightarrow p_\mathcal{L}(\cdot \mid h_\theta(x), i) = p_\mathcal{L}(\cdot \mid \tilde{h}(c), i).
\label{eq:i-align}
\end{equation}
We define \textbf{complete alignment} as the case where $i = \varnothing$ (no instruction). Crucially, complete alignment is a strictly stronger condition:
\begin{equation}
h_\theta(x) \sim_\varnothing \tilde{h}(c) \implies h_\theta(x) \sim_i \tilde{h}(c) \quad \forall i.
\end{equation}
While instruction-tuning only guarantees alignment for instructions within the training set $\mathcal{T}_{\mathrm{train}}$ (\textit{partial alignment}), our instruction-free objective forces $P_\theta$ into global alignment, ensuring \textit{generalization} to any test-time instruction. See Appendix \ref{app:failure-modes} for a detailed analysis of SFT failure modes.

\subsubsection{Information-Capability Bound}
\label{sec:theory-bound}

Let $I(E)$ denote the mutual information~\cite{shannonMI} retained by encoder and $\mathcal{C}(\mathcal{L})$ the response space of the frozen LLM. The reachable performance of an alignment-only system is qualitatively bounded by:
\begin{equation}
    \mathrm{Performance} \le \min \bigl( I(E), \mathcal{C}(\mathcal{L}) \bigr).
    \label{eq:reach-bound}
\end{equation}
By freezing $\mathcal{L}$, we preserve the full breadth of $\mathcal{C}(\mathcal{L})$ and avoid the catastrophic forgetting~\cite{luo2023empirical} typical of task-specific tuning~\cite{li2021prefixtuning, hu2022lora}. The projector's role is thus to minimize the ``slack'' on the information side, ensuring that $h_\theta(x)$ conveys sufficient entropy to saturate the LLM's capacity $\mathcal{C}(\mathcal{L})$.  
However, the extent to which this theoretical ceiling is realized depends on the coverage of the alignment data $\mathcal{D}_{\mathrm{align}}$.
Specifically, the ``slack'' is governed by two factors: (1) \textbf{audio diversity}, which ensures the projector learns to map the entire feature space of $I(E)$, and (2) \textbf{response richness}, which ensures the target $g(c)$ spans a sufficient variety of $\mathcal{L}$'s semantic behaviors. 
Only when $\mathcal{D}_{\mathrm{align}}$ provides sufficient coverage of both modalities can the projector faithfully ``saturate'' the LLM's capability with the encoder's information. Detailed information-theoretical derivations are provided in Appendix \ref{app:mechanism}.

\section{Experiments}
\label{secc:exp}

\subsection{Model Architecture}
\label{sec:model-arch}
Our framework adopts a modular \textit{encoder–projector–LLM} architecture, which follows the established paradigm for multimodal understanding. This modularity allows us to isolate and ablate each component, verifying that our alignment strategy remains robust across different architectural choices.
\vspace{-8mm}
\paragraph{Large Language Model.}
We select Qwen2.5-7B-Instruct~\citep{qwen2025qwen25technicalreport} as our default backbone to leverage its robust open-weight instruction-following capabilities as a representative \textit{universal decoder}.
To demonstrate the cross-generational portability of our alignment recipe, we further evaluate the more recent Qwen3-8B~\citep{yang2025qwen3technicalreport} in our LLM ablation (Sec.~\ref{sec:llm-ablation}).
\vspace{-3mm}
\paragraph{Audio Encoder.}
We select five encoders to evaluate our recipe across three distinct pre-training paradigms:
(1) \textbf{AudioSet-Zipformer}~\cite{tseng2026revisiting}, a Zipformer-based~\cite{yao2024zipformer} model pretrained on AudioSet~\cite{audioset} via multi-tag classification~\cite{mtc}. It represents the discriminative paradigm, where features are optimized for sound-event and acoustic scene recognition; (2) \textbf{Whisper-large-v2}~\citep{whisper}, which is trained with speech transcription supervision, represents the ASR-oriented paradigm; and (3) \textbf{Qwen family} (Qwen2.5-Omni encoder~\cite{xu2025qwen25omnitechnicalreport}, Qwen3-Omni AuT encoder~\cite{Qwen3-Omni} and Qwen3-ASR AuT encoder~\cite{Qwen3-ASR}), representing the joint audio-language pretraining paradigm. Encoder ablation is in Sec.~\ref{sec:encoder-ablation}.
\vspace{-3mm}
\paragraph{Projector.}
To maintain a minimalist and computationally efficient alignment interface, we utilize a simple two-layer MLP as the projector. Its primary role is to bridge the modality gap by merging consecutive encoder frames and mapping them into the LLM's latent space. 
We set the downsampling rate (between 2 and 8) tailored to each encoder's frame rate to maintain a consistent audio-token frequency of 6.25 - 12.5\,Hz. This ensures that the LLM receives a dense yet manageable sequence of auditory information, the impact of which is ablated in Appendix.~\ref{app:ds-rate}.

\subsection{Dataset}
\label{sec:dataset}
We primarily use paired \textit{(audio, caption)} data from \textit{CaptionStew}~\cite{tseng2026revisiting}, a large-scale dataset with 10M samples covering diverse audio sources, including speech, music, and environmental sounds. Specifically, we use its 400K, 1M, and 4M subsets for scaling experiments (See Sec.~\ref{sec:scaling}). Additionally, for captioner-generated data, we adopt captions from \citet{zhou2026uts}, which are produced by the \textit{Qwen3-Omni-Captioner} on the CaptionStew 400K subset.
As CaptionStew contains limited speech data, we additionally incorporate the speech corpora curated by~\citet{shao2026azeros}, including DailyTalk~\cite{dailytalk}, CREMA-D~\cite{cao2014crema}, RAVDESS~\cite{ravdess}, TESS~\cite{tess}, MELD~\cite{poria2018meld}, IEMOCAP~\cite{busso2008iemocap}, VoxCeleb2~\cite{voxceleb2} and CommonVoice-en~\citep{ardila2019common}.
We treat transcripts and paralinguistic annotations as semantic surrogate, functionally equivalent to caption for the corresponding audio.
Dataset statistics are in Appendix~\ref{app:dataset}.

\subsection{Training Configuration}
\label{sec:train-config}
\paragraph{Training Set.}
We construct the training set using the 400K, 1M, and 4M subsets of \textit{CaptionStew}, each augmented with 10\% additional speech data to improve coverage of speech-related task.
Detailed configurations and examples for Self-Generated Data Construction are provided in Appendix~\ref{app:gen-config}. 
\paragraph{Model Setup.}
\label{sec:model-setup}
To ensure distributional consistency, we employ the same LLM instance for both the data expansion phase and the subsequent instruction-free alignment training.
We pair LLMs of comparable scale (e.g., Qwen2.5-7B and Qwen3-8B) to maintain a consistent model size regime.
In all experiments, the audio encoder and the LLM remain \textit{frozen} to preserve their extensive pre-trained priors. Consequently, only the lightweight projector is trained. For example, the \textit{Whisper-large-v2 + Qwen2.5-7B} (projector downsampling rate of 4) involves only 31.2M trainable parameters.
\paragraph{Optimization Details.}
All experiments are conducted on a single node equipped with 8 NVIDIA A100 (40\,GB) GPUs under bf16 mixed precision. 
We optimize the projector with AdamW using a peak learning rate of $1\mathrm{e}{-3}$ and a cosine decay schedule, with gradient clipping at $\ell_2$ norm $1.0$.
Each batch is assembled by a length-bucketed dynamic sampler (30 buckets, drop-last) that caps 
the per-GPU budget at $3{,}000$ feature tokens per step.
We maintain an exponential moving average (EMA) of parameters for evaluation stability.
Training runs for $600$k steps on the CaptionStew 400K subset augmented with 10\% speech data, taking approximately 46 hours on an 8-GPU node.

\subsection{Evaluation Benchmarks}
We evaluate on four audio understanding benchmarks of increasing scope and difficulty. Full evaluation configurations are detailed in Appendix~\ref{app:eval-config}.
\paragraph{MMAU~\cite{MMAU}.}
MMAU contains 10{,}000 audio clips paired with human-annotated QA pairs across speech, environmental sounds, and music, covering 27 tasks that require both information extraction and multi-step reasoning. It is the most widely adopted benchmark in audio understanding community.
\paragraph{MMAR~\cite{ma2025mmar}.}
MMAR comprises 1{,}000 QA triplets sourced from real-world internet videos, covering mixed combinations of speech, sound, and music, with questions organized across four reasoning layers: Signal, Perception, Semantic, and Cultural.
\paragraph{MMSU~\cite{wang2025mmsu}.}
MMSU provides 5{,}000 QA pairs across 47 tasks grounded in linguistic phenomena spanning phonetics, prosody, syntax, semantics, and paralinguistics. MMSU focuses exclusively on \textit{speech}, probing whether models understand how something is said beyond its literal content.
\paragraph{MMAU-Pro~\cite{kumar2025mmau-pro}.}
MMAU-Pro contains 5{,}305 expert-annotated instances across 49 skills, including long-form audio up to 10 minutes. As our most challenging benchmark, we restrict to instances \textit{under 1 minute} and evaluate across three task types: instruction following, open-ended response, and closed QA (encompassing spatial audio reasoning and multi-audio understanding, among others).

\subsection{Main Results}
Table.~\ref{tab:main_results_all} compares our two instantiations against ten open-source LALMs and four proprietary systems on MMAU, MMAR, MMSU, and MMAU-Pro. Both share the frozen Qwen2.5-7B-Instruct backbone, the same projector recipe, and training corpus (CaptionStew 400K with 10\% added speech; 576.8K samples / 1.6K hours), differing only in the audio encoder. The AudioSet-Zipformer variant reaches 80.8\,/\,77.4 on MMAU \textit{Sound}, exceeding the previous open-source best (Audio-Flamingo 3: 79.6\,/\,75.8) trained on $46\times$ more samples and $34\times$ more audio. On MMAU-Pro \textit{Avg.}, it scores 52.8, the highest among the listed open-source LALMs (Audio-Flamingo 3: 51.7; Qwen2.5-Omni: 52.2; Kimi-Audio: 46.6) and on par with GPT-4o-Audio (52.5). It trails Audio-Flamingo 3 (72.4) by 6.1 points and Qwen2.5-Omni (71.0) by 4.7 on MMAU \textit{test} Avg. The Whisper-large-v2 variant covers speech-leaning subsets, reaching 60.4\,/\,60.1 on MMAU \textit{Speech}, 50.6 on MMSU, and 55.4 on MMAU-Pro open-ended response. Ours top every open-source LALM on MMAU-Pro IF (62.9 / 72.6 vs.\ 61.3), reflecting frozen LLM that preserves Qwen2.5-7B-Instruct's instruction-following intact.

Speech is still the weak axis: ALARM reaches 77.2\,/\,73.7 on MMAU \textit{Speech} and 61.3 on MMSU; Qwen2.5-Omni reaches 70.6\,/\,68.9 and 60.6. Both exceed our Whisper variant by 10 to 17 points on these subsets. 
We attribute the gap to the limited speech proportion in the current mix and the \textit{insufficient density} of speech-related information in the original captions, and try to address it with targeted speech-QA SFT in Sec.~\ref{sec:sft}. 
Against proprietary systems, the AudioSet-Zipformer variant lands within 1.1 points of Gemini 2.5 Flash on MMAU \textit{test} Avg.\ (66.3 vs.\ 67.4) and exceeds GPT-4o-Audio (60.8) by 5.5. The largest gap is on MMAR: our 54.3 trails Gemini 2.5 Flash (68.4) by 14.1 points and Audio-Flamingo 3 (58.5) by 4.2.

\providecolor{groupshade}{HTML}{EEEEEE}    
\def\restabcolsep{4pt}                     
\def\restabrowstretch{1.1}                
\def\restabwidth{\textwidth}               

\begin{table*}[t]
\centering
\setlength{\tabcolsep}{\restabcolsep}
\renewcommand{\arraystretch}{\restabrowstretch}
\small

\caption{Comparison on all four audio-understanding benchmarks. MMAU reports \textit{test-mini}\,/\,\textit{test} accuracy across Sound, Music, Speech, and their average; MMAR and MMSU report the overall average; MMAU-Pro
reports instruction following (IF), open-ended response, and overall
average.}
\label{tab:main_results_all}

\resizebox{\restabwidth}{!}{%
\begin{tabular}{@{}l c c c c c c c c c c c@{}}
\toprule
\multirow{2}{*}{\textbf{Model}} &
\multirow{2}{*}{\textbf{Size}} &
\textbf{Training Data} &
\multicolumn{4}{c}{\textbf{MMAU\;(\textit{test-mini}\,/\,\textit{test})}} &
\textbf{MMAR} &
\textbf{MMSU} & 
\multicolumn{3}{c}{\textbf{MMAU-Pro}} \\
\cmidrule(lr){4-7} \cmidrule(lr){8-8} \cmidrule(lr){9-9} \cmidrule(lr){10-12}
& & \# Samples | \# Hours &
Sound  & Music  & Speech & \textit{Avg.} & \textit{Avg.}
& \textit{Avg.} &
IF & Open-ended & \textit{Avg.}
\\
\midrule

\rowcolor{groupshade}
\multicolumn{12}{l}{\textit{\textbf{Proprietary models}}} \\
GPT-4o mini Audio~\cite{openai2024gpt4ocard} &-- &-- & 50.8 / 49.7 & 39.2 / 36.0 & 69.1 / 67.5 & 53.0 / 51.0 & 50.6 & -- &  79.7   & 41.6   & 48.3  \\
GPT-4o-Audio~\cite{openai2024gpt4ocard}    & --   & -- &  64.6 / 63.2 & 56.3 / 49.9 &66.7 / 69.3 & 62.5 / 60.8 & 63.5  & \textbf{56.4} & 82.5 & 43.2 & 52.5 \\
Gemini 2.0 Flash~\cite{google_gemini_2_0_flash} &-- &-- & 71.2	/ 68.9 &	65.3 / 59.3	& 75.1	/ 72.9 & 70.5 / 67.0 & 65.6 & 51.0 & 94.2 & 66.8 & 55.7\\
Gemini 2.5 Flash~\cite{google_gemini_flash} &-- &-- & \textbf{73.3 / 69.5} &	\textbf{65.6 / 69.4} &	\textbf{76.6 / 68.3}	& \textbf{71.8 / 67.4} & \textbf{68.4} & -- & \textbf{95.1} & \textbf{67.5} & \textbf{59.2}\\
\addlinespace[2pt]

\rowcolor{groupshade}
\multicolumn{12}{l}{\textit{\textbf{Open-source/access LALMs}}} \\
SALMONN~\cite{tang2024salmonn}   & 13B  & 2.3M | 4.4K & 41.1 / 42.1 & 37.1 / 37.8 & 26.4 / 28.8 & 34.9 / 36.2 & 33.2 & 30.1 & 38.5 & 33.6 & 39.6 \\
LTU~\cite{gong2024listen} & 7B & 5.6M | -- & 20.4 / 20.7 & 16.0 / 15.7 & 15.9 / 15.3 & 17.4 / 17.2 & 19.2 & 22.6 & -- & -- & -- \\
Qwen2-Audio-Instruct~\cite{Qwen2-Audio}      & 7B & -- | 320K  &  67.3 / 61.2 & 56.3 / 55.7 & 55.3 / 55.4 & 59.6 / 57.4 & 30.0 & 53.3 & -- & -- & --\\
Qwen2.5-Omni~\cite{xu2025qwen25omnitechnicalreport}     & 7B   & -- | --  & 78.1 / 76.8 & 65.9 / 67.3 & 70.6 / 68.9 & 71.5 / 71.0 & 56.7 & 60.6 & \textbf{61.3} & \textbf{52.3} & \textbf{52.2} \\
Audio-Flamingo 2~\cite{af2}   & 3B   & 5.9M | --   &  71.5 / 68.1 & 71.0 / 70.2 & 44.7 / 44.9 & 62.4 / 61.1 & 21.9 & -- & 29.6 & 43.2 & 42.6 \\
Audio-Flamingo 3~\cite{af3}   & 8B   & 26.7M | 54.4K    & \textbf{79.6 / 75.8} & \textbf{74.0 / 74.5} & 66.4 / 67.0 & \textbf{73.3 / 72.4} & \textbf{58.5} & -- & 33.3 & 44.2 & 51.7\\
Kimi-Audio~\cite{kimiteam2025kimiaudiotechnicalreport}    & 7B   & -- | 13.3M    & 75.7 / 70.7	& 66.8 / 65.9	& 62.2 / 56.6 & 68.2 / 64.4 & -- & 59.3 & 42.3 & 34.5 & 46.6\\
ALARM~\cite{alarm} & 4B & 5.5M | 17K & 64.0 / 59.1 & 54.8 / 54.2 &  \textbf{77.2 / 73.7} & 65.3 / 62.4 & 48.7 & \textbf{61.3} & -- & -- & --
\\
\midrule

\textbf{Ours} (AudioSet-Zipformer)    & 7B   & 576.8K | 1.6K & \textbf{80.8 / 77.4} & \textbf{69.8 / 68.6} & 54.1 / 52.8 & \textbf{68.2 / 66.3} & \textbf{54.3} & 47.0 & 62.9 & 50.8 & \textbf{52.8} \\
\textbf{Ours} (Whisper-large-v2)    & 7B   & 576.8K | 1.6K & 75.7 / 72.5 & 62.9 / 61.3 & \textbf{60.4 / 60.1} & 66.3 / 64.6 & 52.3 & \textbf{50.6} & \textbf{72.6} & \textbf{55.4} & 48.4 \\
\bottomrule
\end{tabular}%
}
\end{table*}

\subsection{Analysis}

\subsubsection{Effect of Audio Encoder}
\label{sec:encoder-ablation}

Table.~\ref{tab:encoder_ablation} sweeps five encoders across three pretraining paradigms with the LLM, training corpus, and projector recipe held fixed. The discriminative AudioSet-Zipformer leads on MMAU \textit{Avg.} (68.2), MMAR (54.3), and MMAU-Pro \textit{Avg.} (52.8), with the largest leads on Sound (80.8) and Music (69.8), matching AudioSet~\cite{audioset} supervision targets. 
Whisper-large-v2 excels on speech-leaning subsets (MMAU \textit{Speech} 60.4, MMSU 50.6), aligning with its ASR pretraining. The joint audio-language encoders (Qwen2.5-Omni encoder, Qwen3-Omni AuT encoder) trail AudioSet-Zipformer by 7-13 points on MMAU \textit{Avg.}, suggesting that representations co-tuned to a specific decoder \textit{transfer only partially} under our frozen-LLM, projector-only recipe. Across the five encoders, the per-benchmark winner \textit{tracks the axis emphasized during pretraining}, consistent with the $\min(I(E), \mathcal{C}(\mathcal{L}))$ bound (Sec.~\ref{sec:theory-bound}).
With $\mathcal{L}$ fixed, the encoder selects which benchmarks the recipe can reach.

\providecolor{lightblue}{HTML}{E1F1FF}

\begin{table*}[t]
\centering
\setlength{\tabcolsep}{\restabcolsep}
\renewcommand{\arraystretch}{\restabrowstretch}
\small

\caption{Effect of audio encoder. The LLM (Qwen2.5-7B-Instruct) and training data
(CaptionStew 400K + 10\% speech) are held fixed;
only the audio encoder is swapped, with the projector downsampling rate $r$
adjusted per encoder to maintain a post-projector token rate of 6.25--12.5\,Hz.}
\label{tab:encoder_ablation}
\vspace{3pt}

\resizebox{\restabwidth}{!}{%
\begin{tabular}{@{}l c c c c c c c c c c c@{}}
\toprule
\multirow{2}{*}{\textbf{Audio Encoder}} &
\multirow{2}{*}{\textbf{Size}} &
\multirow{2}{*}{\textbf{Down.\ $r$}} &
\multicolumn{4}{c}{\textbf{MMAU\;(\textit{test-mini})}} &
\textbf{MMAR} &
\textbf{MMSU} & 
\multicolumn{3}{c}{\textbf{MMAU-Pro}} \\
\cmidrule(lr){4-7} \cmidrule(lr){8-8} \cmidrule(lr){9-9} \cmidrule(lr){10-12}
& & &
Sound  & Music  & Speech & \textit{Avg.} & \textit{Avg.}
& \textit{Avg.} &
IF & Open-ended & \textit{Avg.}
\\
\midrule

\rowcolor{lightblue}
\multicolumn{12}{l}{\textit{\textbf{Discriminative pretraining}}} \\
AudioSet-Zipformer        & 7B   & 4 & \textbf{80.78} & \textbf{69.76} & 54.05 & \textbf{68.20} & \textbf{54.30} & 47.00 & 62.91 & 50.84 & \textbf{52.82}  \\
\addlinespace[2pt]

\rowcolor{lightblue}
\multicolumn{12}{l}{\textit{\textbf{ASR-supervised pretraining}}} \\
Whisper-large-v2          & 7B   & 4 & 75.68 & 62.87 & \textbf{60.36} & 66.30 & 52.30 & \textbf{50.61} & \textbf{72.57} & 55.39 & 48.40\\
Qwen3-ASR AuT encoder    & 7B   & 2   & 63.36 & 51.80 & 55.15 & 56.80 & 47.20 & 45.15 & 59.43 & 52.12 & 41.92\\
\addlinespace[2pt]

\rowcolor{lightblue}
\multicolumn{12}{l}{\textit{\textbf{Joint audio-language pretraining}}} \\
Qwen2.5-Omni audio encoder      & 7B  & 4    & 68.77 & 50.9 & 45.65 & 55.10 & 48.30 & 43.31 & 67.74 & 53.77 & 45.24\\
Qwen3-Omni AuT encoder    & 7B   & 2  & 67.87 & 60.78 & 54.95 & 61.20 & 48.40 & 45.81 & 67.74 & \textbf{60.59} & 47.32\\
\bottomrule
\end{tabular}%
}
\vspace{-3mm}
\end{table*}

\vspace{-3pt}
\subsubsection{Effect of LLM Backbone}
\label{sec:llm-ablation}
\vspace{-3pt}
Table.~\ref{tab:llm_ablation} swaps the LLM with the encoder (Qwen3-Omni AuT), training corpus, and projector recipe held fixed. Under the matched setting (same LLM for $\mathcal{S}$ generation and alignment), Qwen2.5-7B-Instruct and Qwen3-8B reach the same band: MMAU \textit{Avg.} 61.2 vs.\ 60.6, MMAR 48.4 vs.\ 49.8, MMAU-Pro \textit{Avg.} 47.3 vs.\ 48.3. The cross-generational match supports porting the recipe to newer LLM releases through projector retraining alone. When the $\mathcal{S}$ generator and the alignment LLM differ (Qwen2.5-7B-Instruct generating data for Qwen3-8B alignment), MMAU \textit{Avg.} drops 5.9 points (60.6 $\to$ 54.7) and MMAU-Pro \textit{Avg.} drops 7.8 (48.3 $\to$ 40.5). The drop confirms the matched-generator setup of Sec.~\ref{sec:model-setup} is a precondition for portability, not a stylistic choice.

\begin{table*}[t]
\centering
\setlength{\tabcolsep}{\restabcolsep}
\renewcommand{\arraystretch}{\restabrowstretch}
\small

\caption{Effect of LLM backbone. The audio encoder, training data, and projector
configuration are held fixed; only the LLM is swapped.
We compare matched and mismatched settings, where the same or different LLM is used for response generation ($\mathcal{S}$ generator) and alignment training.}
\label{tab:llm_ablation}

\resizebox{\restabwidth}{!}{%
\begin{tabular}{@{}l c l c c c c c c c c c@{}}
\toprule
\multirow{2}{*}{\textbf{LLM Backbone}} &
\multirow{2}{*}{\textbf{Size}} &
\multirow{2}{*}{\textbf{$\mathcal{S}$ Generator}} &
\multicolumn{4}{c}{\textbf{MMAU\;(\textit{test-mini})}} &
\textbf{MMAR} &
\textbf{MMSU} & 
\multicolumn{3}{c}{\textbf{MMAU-Pro}} \\
\cmidrule(lr){4-7} \cmidrule(lr){8-8} \cmidrule(lr){9-9} \cmidrule(lr){10-12}
& & &
Sound  & Music  & Speech & \textit{Avg.} & \textit{Avg.}
& \textit{Avg.} &
IF & Open-ended & \textit{Avg.}
\\
\midrule

\rowcolor{lightblue}
\multicolumn{12}{l}{\textit{\textbf{Qwen2.5 family}}} \\
Qwen2.5-7B-Instruct       & 7B   & Qwen2.5-7B-Instruct & \textbf{67.87} & 60.78 & \textbf{54.95} & \textbf{61.20} & 48.40 & 45.81 & \textbf{67.74} & \textbf{60.59} & 47.32 \\
\addlinespace[2pt]

\rowcolor{lightblue}
\multicolumn{12}{l}{\textit{\textbf{Qwen3 family}}} \\
Qwen3-8B                  & 8B   & Qwen2.5-7B-Instruct  & 63.44	& 55.18 & 45.43 & 54.70 & 48.10 & 45.84 & 61.09 & 50.32 & 40.54\\
Qwen3-8B                  & 8B   & Qwen3-8B  & 66.77 & \textbf{61.80}	& 53.25 & 60.60 & \textbf{49.80} & \textbf{46.86} & \textbf{67.74} & 54.81 & \textbf{48.31} \\
\bottomrule
\end{tabular}%
}
\vspace{-5mm}
\end{table*}

\begin{figure*}[t]
\label{fig:exp1}
    \centering
    \hspace*{-0.05\textwidth}
\resizebox{1.1\textwidth}{!}{\includegraphics{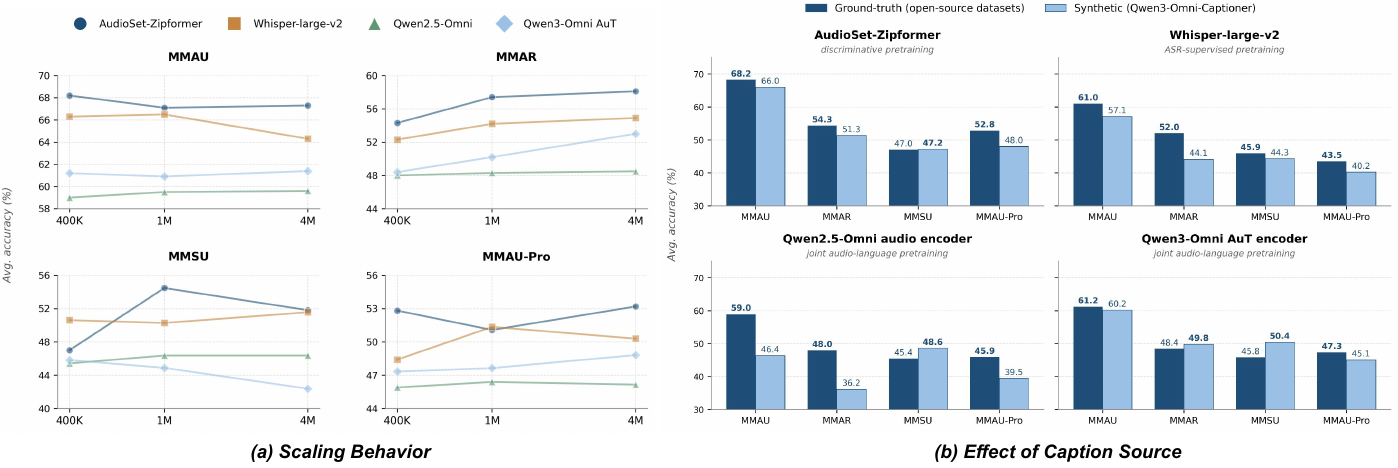}}
    \caption{(a) Scaling behavior of four audio encoders across increasing dataset sizes on four benchmarks. (b) Effect of caption source: comparison between ground-truth and synthetic captions.}
    \vspace{-3mm}
\end{figure*}

\subsubsection{Scaling Behavior}
\label{sec:scaling}

Fig.2 (a) shows scaling behavior from CaptionStew 400K to 4M subsets across four encoders. The four panels exhibit distinct trends rather than a shared scaling curve, indicating that the effect of data scaling depends on the type of task being evaluated. 
MMAU, which probes closed-set audio understanding through multiple-choice questions, is flat for every encoder across the full sweep, with no monotone gain on any line. 
MMAR, which rewards reasoning over multi-event audio, rises on three of the four encoders, with the steepest gain on the newest encoder (Qwen3-Omni). 
MMSU and the MMAU-Pro average drift without a shared direction; per-encoder noise dominates the trend. 
Overall, scaling alignment data primarily benefits tasks that require \textit{richer, open-ended reasoning}, while having little effect on closed-set recognition. This is because recognition performance is constrained by the encoder representations, which the projector cannot fundamentally alter.
Once a minimal amount of data enables alignment, additional data yields diminishing returns, as performance is bounded by the combined capacity of the encoder and LLM.
This observation is consistent with the Information-Capability Bound (Sec.~\ref{sec:theory-bound}), where the limit is bounded by encoder informativeness and LLM capability rather than alignment data. Detailed breakdowns are provided in Appendix~\ref{app:scaling-contd}.

\subsubsection{Effect of Caption Source}
\label{subsubec:caption}
Fig.2 (b) compares the two sources for $\mathcal{S}$
across four encoders. Ground-truth captions hold a small but consistent
advantage for AudioSet-Zipformer, Whisper-large-v2, and Qwen2.5-Omni, with the
largest gap (12.6 points on MMAU) under Qwen2.5-Omni. The advantage collapses
for Qwen3-Omni AuT: synthetic captions match ground truth on MMAU and MMAU-Pro
and surpass it on MMAR (+1.4) and MMSU (+4.6). This suggests that synthetic captions provide denser and more complete semantic information, especially beneficial for speech-centric tasks like MMSU that require richer linguistic and paralinguistic cues.
Overall, synthetic captions can be competitive with strong encoders, but the consistent strength of ground-truth captions indicates that our recipe \textit{does not require dense synthetic descriptions for every sample}. Instead, simple open-source captions are sufficient when aggregated over many diverse samples, where information emerges from dataset-level coverage rather than exhaustive per-sample annotation.
Per-subset breakdowns are in
Appendix~\ref{app:caption-contd}.

\vspace{-3pt}
\subsubsection{Targeted SFT for the Speech Axis}
\label{sec:sft}
\vspace{-3pt}
\begin{wraptable}{r}{0.45\textwidth}
\centering
\vspace{-7mm}
\setlength{\tabcolsep}{4pt}
\renewcommand{\arraystretch}{1.05}
\footnotesize
\caption{Effect of targeted speech-QA SFT on the \textit{(Whisper-large-v2, Qwen2.5-7B)} recipe.}
\label{tab:sft_ablation}
\vspace{2pt}
\begin{tabular}{@{}l c c c c@{}}
\toprule
\multirow{2}{*}{\textbf{Setting}} &
\multicolumn{4}{c}{\textbf{MMAU\;(\textit{test-mini})}} \\
\cmidrule(lr){2-5}
& Sound & Music & Speech & \textit{Avg.} \\
\midrule
Alignment Only      & \textbf{75.68} & \textbf{62.87} & 60.36          & \textbf{66.30} \\
Alignment $\to$ SFT & 70.87          & 58.68          & \textbf{65.47} & 65.00          \\
\bottomrule
\end{tabular}
\vspace{-3mm}
\end{wraptable}

The Whisper variant lags on speech-leaning subsets (Sec.~\ref{sec:encoder-ablation}), which we attributed to the limited speech proportion and information density in training data. To test whether the gap is data-bound rather than recipe-bound, we fine-tune the aligned projector for an additional 60K steps on speech-QA data from \textbf{AudioSkills-XL}~\cite{af3}, keeping the encoder and LLM frozen. Detailed dataset statistics are provided in Appendix~\ref{app:dataset}.
Table.~\ref{tab:sft_ablation} reports the result on MMAU. SFT lifts \textit{Speech} by 5.1 points (60.36 $\to$ 65.47), with the gain concentrated on format-aligned tasks (Multi-Speaker Role Mapping: 55.6 $\to$ 92.6; Phonemic Stress Pattern Analysis: 47.2 $\to$ 50.9). The expected trade-off appears on the non-targeted axes (Sound $-4.8$, Music $-4.2$), as the late SFT phase pulls the projector toward the QA distribution at the cost of the broader caption-aligned representation. The speech gap is thus closable on demand, while the instruction-free recipe remains a strong default for general-purpose audio understanding.

\section{Limitation and Conclusion}
\label{sec:con}
\vspace{-3pt}
We introduce an Instruction-Free Alignment-Only training recipe for LALMs. 
By freezing both the audio encoder and LLM and training only a lightweight projector, our method achieves competitive performance without extensive post-training. 
Extensive ablations demonstrate that competitive MLLM can emerge from alignment alone, simplifying multimodal LLM construction to a lightweight projector-training task.
Our work also has several limitations. The empirical study is limited to the audio domain and the 7B scale, leaving validation on larger models and other modalities for future work.
Moreover, the achievable performance ceiling is fundamentally bounded by the base model: alignment cannot recover missing encoder information or induce new LLM capabilities. 
Finally, caption/response-based supervision primarily captures semantics information, while some richer super-semantic signals may require specially QA supervision or stronger representations.

\medskip

\newpage
{
\small
\bibliographystyle{unsrtnat}
\bibliography{reference}
}


\appendix
\section{Detailed Analysis of SFT Failure Modes}
\label{app:failure-modes}

While SFT is effective for optimizing specific tasks, it introduces two phenomena that hinder the development of a general-purpose audio-language interface.

\paragraph{Instruction-Induced Feature Collapse.}
During SFT, the gradient $\nabla_\theta \mathcal{L}$ is conditioned on a specific task $i$ (e.g., Automatic Speech Recognition~\cite{whisper}). For such a task, audio features irrelevant to the transcript, such as emotional prosody, background acoustics, or speaker identity, contribute nothing to the loss reduction. Consequently, the projector $P_\theta$ learns to act as an \textit{information filter}, actively suppressing these ``irrelevant'' dimensions. This collapse is irreversible; once these features are pruned during training, the model cannot recover them for unseen tasks at test time (e.g., being asked to analyze the speaker's mood).

\paragraph{Modality Neglect (The Shortcut Problem).} 
High-capacity LLMs possess powerful linguistic priors. When an instruction $i$ and the target response $y$ are highly correlated, the model often discovers a ``shortcut'' by modeling $p(y \mid i)$ directly~\cite{8100153, lin2024revisiting}, effectively bypassing the audio prefix $h_\theta(x)$. This leads to a model that is ``dishonest'' to the sensory input; it learns to follow task templates and linguistic patterns rather than grounding its generation in the actual auditory evidence.

\section{The Information-Theoretical Mechanism of Global Alignment}
\label{app:mechanism}

The failure modes described in Appendix \ref{app:failure-modes} can be interpreted through the lens of the Information Bottleneck (IB) principle \citep{tishby99information}. In this section, we provide a deeper analysis of how our \textbf{Instruction-Free Alignment-Only} framework achieves the theoretical bounds established in Sec.~\ref{sec:theory-bound}.

\paragraph{Maximizing Mutual Information via High-Entropy Targets.}
In a standard SFT objective, the model minimizes $\mathcal{L}_{\mathrm{SFT}} = -\log p(y \mid x, i)$. According to the IB principle, the optimization process encourages the intermediate representation $h_\theta(x)$ to compress and discard all information in $x$ that is not strictly necessary for predicting the task-specific label $y$. 
When $i$ is a narrow task (e.g., classification or event detection), $h_\theta(x)$ effectively becomes a sparse ``attribute-specific filter,'' discarding rich acoustic textures, temporal dynamics, and spatial cues not required for that discrete label.
In contrast, our instruction-free objective $\mathcal{L}_{\mathrm{align}} = -\log p(g(c) \mid h_\theta(x))$ utilizes an LLM-expanded response $g(c)$ as the target. Since $g(c)$ is a high-entropy, descriptive response covering multiple facets of the audio (semantics, acoustics, and environment), the projector is forced to maximize the Mutual Information~\cite{hjelm2018learning} $I(h_\theta(x); x)$. This ensures that the ``slack'' mentioned in Eq.~\ref{eq:reach-bound} is minimized, as the projector must preserve nearly all recoverable information from the encoder to satisfy the LLM's demand for descriptive detail.

\paragraph{Distributional Matching on the Latent Manifold.}
The \textit{universal decoder} hypothesis implies that the LLM's input space constitutes a structured semantic manifold. The textual representation $\tilde{h}(c)$ of a faithful caption already occupies the ``ground-truth'' coordinates on this manifold. Our complete alignment objective ($i = \varnothing$ in Eq.~\ref{eq:i-align}) is essentially a form of distributional matching. By forcing $h_\theta(x)$ to elicit the same response distribution as $\tilde{h}(c)$ without any task-specific guidance, we anchor the audio prefix to the same manifold coordinates as the text. Because the LLM is frozen, this manifold provides a stable reference frame. Once global alignment is achieved, any subsequent test-time instruction $i$ (which acts as a directional perturbation on the manifold) will guide the LLM toward the correct output, leveraging its pre-trained reasoning paths without further adaptation.

\paragraph{Strategic Advantages of a Frozen Backbone.}
While updating $\mathcal{L}$ during alignment is technically feasible, maintaining a frozen backbone serves as a strategic choice for both optimization stability and parameter efficiency. From an optimization perspective, a frozen LLM provides a static semantic anchor. If $\mathcal{L}$ were updated on a relatively small alignment corpus, its high-dimensional manifold might undergo task-specific warping. By keeping $\mathcal{L}$ fixed, the projector's learning objective is simplified to finding a consistent mapping into a stable, high-capacity semantic space.

Furthermore, this approach ensures that the model preserves the universal reasoning priors acquired during massive-scale pre-training. Since alignment data is typically much smaller in volume than the original LLM training sets, freezing the backbone prevents the catastrophic forgetting of general-purpose capabilities, ensuring the model remains a universal decoder rather than a task-specialized tool. This design allows for competitive performance with a minimal computational footprint and rapid convergence.

\section{Zipformer Model}

In experiments on the effect of audio encoder, we adopt the Zipformer-M architecture~\cite{yao2024zipformer} as one of our audio encoder, chosen for its memory efficiency on long sequences and strong performance across audio tasks. 
The model architecture employs a U-Net-inspired design with six Transformer stages that process sequences at multiple temporal resolutions, capturing both fine- and coarse-grained temporal patterns.
Each stage operates at progressively decreasing and then increasing frame rates (50, 25, 12.5, 6.25, 12.5, and 25 Hz), with residual and upsampling connections between stages.

It implements the original 2,2,3,4,3,2 block configuration, where each number indicates the number of blocks per stage, , and fuse the outputs at 25 Hz to produce 768-dimensional frame-level embeddings.
Architectural enhancements such as BiasNorm, Swoosh activations, and compatibility with the ScaledAdam optimizer improve training stability and convergence on long sequences.
While Zipformer was initially designed for automatic speech recognition, prior works~\cite{tseng2026revisiting, zhou2026uts} have demonstrated its effectiveness as a general audio encoder across diverse domains.

Following these studies, we pretrained the Zipformer-M on AudioSet~\cite{audioset} using multi-tag classification over 527 classes by adding a linear classifier after the encoder. The resulting embeddings are then used as the input representation for alignment-only training.

\vspace{-5pt}
\section{Source Datasets for CaptionStew}

We utilize subsets from \textit{CaptionStew}~\citep{tseng2026revisiting}, a composite dataset aggregating 8 open-source collections to mitigate data scarcity and enhance diversity in audio pre-training.
As outlined in Table.~\ref{tab:source}, the data spans diverse acoustic domains, including environmental sounds, music, and expressive speech.

\begin{table*}[th]
\caption{Overview of the public datasets constituting CaptionStew. The table summarizes their scale, domain coverage, audio sources, and diverse captioning pipelines (from human annotation to LLM generation).}
\vspace{5pt}
\label{tab:source}
\resizebox{\textwidth}{!}{
\begin{tabular}{llllll}
\toprule
\textbf{Dataset}                                                      & \textbf{\#audio/\#cap}                                                                 & \textbf{Domain}                                       & \textbf{Audio source}                                                                                 & \textbf{Caption style}                                                                                                                                      & \textbf{Caption generation pipeline}                                                                                                                          \\
\midrule
\begin{tabular}[c]{@{}l@{}}AudioCaps\\ \end{tabular} & 46k/46k                                                                       & \begin{tabular}[c]{@{}l@{}}general (environmental,\\  human/animal sounds)\end{tabular} & \begin{tabular}[c]{@{}l@{}}AudioSet\\ \end{tabular}                                                                                     & \begin{tabular}[c]{@{}l@{}}Human-annotated, short description\end{tabular}                                                                                                        & crowdsourced                                                                                                                                         \\[1.0em] \midrule
\begin{tabular}[c]{@{}l@{}}Clotho \end{tabular}                                                       & 5k/25k                                                                        & environmental sounds                         & FreeSound                                                                                    & \begin{tabular}[c]{@{}l@{}}Human-annotated, short description\end{tabular}                                                                                       & crowdsourced                                                                                                                                         \\ \midrule
\begin{tabular}[c]{@{}l@{}}MusicCaps\\ \end{tabular}                                                   & 3k/3k                                                                         & music                                        & \begin{tabular}[c]{@{}l@{}}AudioSet\\ \end{tabular}                                                                         & \begin{tabular}[c]{@{}l@{}}Expert musician-written,\\multi-sentence, fine-grained description\end{tabular}                                                                                  & expert curation                                                                                                                                      \\[1.0em] \midrule
\begin{tabular}[c]{@{}l@{}}WavCaps\\ \end{tabular}                                                       & 400k/400k                                                                     & \begin{tabular}[c]{@{}l@{}}general (environmental,\\  human/animal sounds)\end{tabular} & \begin{tabular}[c]{@{}l@{}}AudioSet \\ BBC Sound Effect\\ FreeSound\\ SoundBible\end{tabular} & LLM-refined captions                                                                                 & \begin{tabular}[c]{@{}l@{}}three-stage pipeline:\\web-crawled raw descriptions\\ $\rightarrow$ ChatGPT rewrite $\rightarrow$ filtering\end{tabular} \\[1.0em] \midrule
\begin{tabular}[c]{@{}l@{}}AudioSetCaps\\ \end{tabular}                                                  & \begin{tabular}[c]{@{}l@{}}1.9M/1.9M\\ 4.0M/4.0M\\ 182k/182k\end{tabular}     & \begin{tabular}[c]{@{}l@{}}general (environmental,\\  human/animal sounds)\end{tabular} & \begin{tabular}[c]{@{}l@{}}AudioSet \\ YouTube8M\\  VggSound\\ \end{tabular}                      & \begin{tabular}[c]{@{}l@{}}LLM-generated, detailed,\\multi-sentence description \end{tabular}                                                                                                   & \begin{tabular}[c]{@{}l@{}}three-stage pipeline:\\LALM attribute extraction \\ $\rightarrow$ LLM captioning \\$\rightarrow$ CLAP-based filtering\end{tabular}   \\[1.0em] \midrule
\begin{tabular}[c]{@{}l@{}}FusionAudio\\ \end{tabular}                                    & 1.2M/1.2M                                                                     & \begin{tabular}[c]{@{}l@{}}general (environmental,\\  human/animal sounds)\end{tabular} & \begin{tabular}[c]{@{}l@{}}AudioSet\\ \end{tabular}        & \begin{tabular}[c]{@{}l@{}}LLM-augmented, multi-sentence,\\visual-enhanced description \end{tabular}                                                                                                    & \begin{tabular}[c]{@{}l@{}}multimodal context fusion \\(audio, visual, metadata)\\ + LLM captioning\end{tabular}                                       \\[1.0em] \midrule
\begin{tabular}[c]{@{}l@{}}JamendoMaxCap\\  \end{tabular}           & 360k/1.8M                                                                     & music                                        & Jamendo Platform                                                                                     & \begin{tabular}[c]{@{}l@{}}LLM-augmented, multi-sentence,\\fine-grained music description \end{tabular}                                                                                              & \begin{tabular}[c]{@{}l@{}}retrieval-based\\metadata imputation\\ + LLM captioning\end{tabular}                                                       \\[1.0em] \midrule
\begin{tabular}[c]{@{}l@{}}ParaSpeechCaps\\ \end{tabular}             & \begin{tabular}[c]{@{}l@{}}116k/116k (base)\\ 924k/924k (scaled)\end{tabular} & expressive speech                            & \begin{tabular}[c]{@{}l@{}}VoxCeleb1\\  VoxCeleb2\\  EARS\\  Expresso\\  Emilia\\ \end{tabular}     & \begin{tabular}[c]{@{}l@{}}Human-annotated/LLM-augmented,\\speaking-style description\end{tabular} & \begin{tabular}[c]{@{}l@{}}crowdsourced / \\ retrieval-based\\metadata imputation\\ + LALM captioning\end{tabular}    \\
\bottomrule
\end{tabular}
}
\end{table*}

\vspace{-5pt}
\section{Dataset Statistics}
\label{app:dataset}

Table.~\ref{tab:dataset_stats} lists every corpus used in this work. Instruction-Free data pairs each clip with a caption (or transcript / paralinguistic annotation as caption surrogate) and trains the projector under the instruction-free recipe. The \textit{speech corpora} block is the 10\% speech mixture appended to each CaptionStew subset to lift speech coverage. The QA block is used only for the targeted SFT in Sec.~\ref{sec:sft}.

\providecolor{lightgreen}{HTML}{E8F5E9}
\begin{table*}[t]
\centering
\setlength{\tabcolsep}{\restabcolsep}
\renewcommand{\arraystretch}{\restabrowstretch}
\small

\caption{Dataset statistics. \textit{Audio Coverage} marks the audio modalities a corpus contributes: Sp = speech, So = environmental sound, Mu = music. \textit{Hours} is the audio duration actually consumed in our experiments. \textit{Purpose} is either Instruction-Free Alignment (IFA) or speech-QA SFT.}
\label{tab:dataset_stats}
\vspace{3pt}
\resizebox{0.98\textwidth}{!}{%
\begin{tabular}{@{}l l c r c@{}}
\toprule
\textbf{Dataset} & \textbf{Caption Source} & \textbf{Audio Coverage} & \textbf{Hours (h)} & \textbf{Purpose} \\
\midrule

\rowcolor{groupshade}
\multicolumn{5}{l}{\textit{\textbf{Caption corpora (CaptionStew subsets)}}} \\
CaptionStew-400K~\cite{tseng2026revisiting}            & open-source caption  & Sp\,/\,So\,/\,Mu & 1{,}351  & IFA \\
CaptionStew-1M~\cite{tseng2026revisiting}              & open-source caption  & Sp\,/\,So\,/\,Mu & 3{,}475  & IFA \\
CaptionStew-4M~\cite{tseng2026revisiting}              & open-source caption  & Sp\,/\,So\,/\,Mu & 13{,}934 & IFA \\
Qwen3-Omni-Captioner on CS-400K~\cite{zhou2026uts}     & captioner-generated       & Sp\,/\,So\,/\,Mu & 1{,}351  & IFA \\
\addlinespace[2pt]

\rowcolor{groupshade}
\multicolumn{5}{l}{\textit{\textbf{Speech corpora (10\% mixture, transcripts / paralinguistic labels as caption surrogate)}}} \\
DailyTalk~\cite{dailytalk}              & transcript, emotion         & Sp & 21 & IFA \\
CREMA-D~\cite{cao2014crema}             & transcript, gender, age, emotion  & Sp & 5 & IFA \\
RAVDESS~\cite{ravdess}                  & transcript, gender, emotion  & Sp & 1 & IFA \\
TESS~\cite{tess}                        & transcript, gender, age, emotion  & Sp & 1 & IFA \\
MELD~\cite{poria2018meld}               & transcript, gender, emotion  & Sp & 8 & IFA \\
IEMOCAP~\cite{busso2008iemocap}         & transcript, gender, emotion  & Sp & 9 & IFA \\
VoxCeleb2~\cite{voxceleb2}              & transcript, gender & Sp & 2,026 & IFA \\
CommonVoice-en~\cite{ardila2019common}  & transcript, gender, age           & Sp & 1,199 & IFA \\
\addlinespace[2pt]

\rowcolor{groupshade}
\multicolumn{5}{l}{\textit{\textbf{QA corpus (targeted speech SFT)}}} \\
\multirow{2}{*}{AudioSkills-XL~\cite{af3} (speech subset,} & \multirow{2}{*}{QA pairs} & \multirow{2}{*}{Sp} & \multirow{2}{*}{234} & \multirow{2}{*}{SFT} \\
\multirow{2}{*}{incl. VoxCeleb2, GigaSpeech)} & & & & \\
\addlinespace[4pt]
\bottomrule
\end{tabular}%
}
\end{table*}

\vspace{-5pt}
\section{Configuration for Response Generation}
\label{app:gen-config}

\definecolor{msgbg}{HTML}{F7F7F7}
\definecolor{msgframe}{HTML}{8A8A8A}
\newtcblisting{msgbox}[2][]{%
  enhanced, breakable,
  colback=msgbg, colframe=msgframe,
  fonttitle=\bfseries\small, coltitle=black,
  colbacktitle=msgbg!70!white,
  title=#2,
  boxrule=0.4pt, arc=2pt,
  left=4pt, right=4pt, top=2pt, bottom=2pt,
  listing only,
  listing options={
    basicstyle=\footnotesize\ttfamily,
    breaklines=true, breakatwhitespace=true,
    columns=fullflexible,
    keepspaces=true,
    showstringspaces=false,
  },
  #1
}

This section describes the response-generation stage of Self-Generated Data Construction (Sec.~\ref{sec:data-construction}), where an LLM expander $g$ transforms each caption $c$ into a free-form response $r = g(c)$ that serves as the training target. 
Examples of audio, captions, and their corresponding generated responses can be found in Appendix~\ref{app:sample}.

\paragraph{Expander and prompt.}
We use Qwen2.5-7B-Instruct (or Qwen3-8B) as $g$. A short system prompt frames the model as the listener; the caption fills the user turn unmodified. Captions whose tokenized length exceeds $7{,}500$ are tail-truncated before templating, which leaves enough room inside the $8{,}192$-token context window for the response.

\begin{msgbox}{Response generation --- caption expansion}
message = [
    {"role": "system", "content": "You are an AI assistant directly hearing this audio. Respond as if you heard it yourself."},
    {"role": "user",   "content": caption},
]
\end{msgbox}

\paragraph{Decoding.}
We sample with temperature $0.6$, top-$p$ $0.9$, top-$k$ $20$, and $\text{max\_new\_tokens}=512$. Outputs that hit the token budget receive a \texttt{<|truncated|>} suffix so they can be filtered before entering $\mathcal{D}_{\mathrm{align}}$.

\paragraph{Serving.}
Generation runs under vLLM on four NVIDIA A100 (40\,GB) GPUs in bf16, with tensor-parallel size $4$, GPU-memory utilization $0.9$, and a fixed seed of $1234$. CPU producers apply the chat template in parallel and feed a single vLLM engine that batches up to $\text{max\_num\_seqs}=32$ prompts at a time.

\section{Message Templates and Decoding}
\label{app:eval-config}

This section documents the message templates and decoding settings used at training and evaluation. 

\paragraph{Training templates.}
Alignment-Only training stays instruction-free: the user turn carries only the audio token, and the assistant turn carries the targeted response. The optional speech-QA SFT (Sec.~\ref{sec:sft}) prepends the question to the user turn but keeps the same two-turn shape. Neither template uses a system prompt.

\begin{msgbox}{Training --- instruction-free alignment}
message = [
    {"role": "user",      "content": f"{DEFAULT_AUDIO_TOKEN}"},
    {"role": "assistant", "content": response},
]
\end{msgbox}

\begin{msgbox}{Training --- QA SFT}
message = [
    {"role": "user",      "content": f"{DEFAULT_AUDIO_TOKEN}{question}"},
    {"role": "assistant", "content": answer},
]
\end{msgbox}

\paragraph{Evaluation templates.}
Two prompt variants cover the four benchmarks. Closed-set items (MMAU, MMAR, MMSU, and the closed-QA split of MMAU-Pro) use the multiple-choice template, which forces the model to return the answer alone. Open-ended items (instruction-following and the open-ended splits of MMAU-Pro) use the second template, which only asks for concision. Both add a short system prompt and leave the assistant turn empty for the model to fill.

\begin{msgbox}{Evaluation --- multiple-choice QA}
message = [
    {"role": "system",    "content": "You are an audio understanding assistant. Listen to the audio and answer the question directly. Give only the answer, no explanation."},
    {"role": "user",      "content": f"{DEFAULT_AUDIO_TOKEN}{question}"},
    {"role": "assistant", "content": ""},
]
\end{msgbox}

\begin{msgbox}{Evaluation --- open-ended response}
message = [
    {"role": "system",    "content": "You are an audio understanding assistant. Listen to the audio and answer the question concisely."},
    {"role": "user",      "content": f"{DEFAULT_AUDIO_TOKEN}{question}"},
    {"role": "assistant", "content": ""},
]
\end{msgbox}

\paragraph{Decoding.}
We decode with beam search ($\text{num\_beams}=4$), $\text{max\_tokens}=256$, and a per-batch $\text{max\_duration}=500$ audio seconds. Inference runs on a single NVIDIA A100 (40\,GB) under bf16, using the projector's exponential moving average (EMA) weights from training (Sec.~\ref{sec:model-setup}).

\section{Projector Downsampling Rate}
\label{app:ds-rate}
The projector downsamples encoder output by a factor $r$ along the time axis before passing tokens to the LLM, so $r$ controls audio-token density at the LLM input. Native encoder frame rates differ: 50\,Hz for Whisper-large-v2, 25\,Hz for AudioSet-Zipformer and Qwen2.5-Omni audio encoder, 12.5\,Hz for the Qwen3-Omni AuT encoder. Our default in Sec.~\ref{sec:encoder-ablation} picks the $r$ that lands the audio-token frequency in 6.25--12.5\,Hz. Table.~\ref{tab:ds_rate} sweeps $r$ around that band per encoder.
One principle organizes the sweep: audio-token frequency governs performance, not $r$ itself. Every encoder's best setting lands inside the 6.25-12.5\,Hz band. Rows outside the band degrade, and they degrade asymmetrically. Inside the 6.25-12.5\,Hz band, the residual choice is benchmark-dependent. Speech and instruction-following favor the dense end; sound and music favor the sparse end.

Whisper-large-v2 makes this clearest. At $r=4$ (12.5\,Hz) it reaches MMAU 66.30 and MMSU 50.61, the strongest result in the section. Halving $r$ doubles the audio-token frequency to 25\,Hz, and MMAU collapses to 46.10. The damage is uneven across modalities: speech drops 29 points (60.36 to 31.53), music drops 20 (62.87 to 42.81), and sound drops only 12 (75.68 to 63.96). Doubling $r$ thins to 6.25\,Hz and costs 5 MMAU points; quadrupling to 3.125\,Hz costs another 2. Density past the band breaks sharply, content-rich modalities first; density below the band costs gradually.

Joint-pretrained encoders show the same shape. The Qwen2.5-Omni audio encoder peaks at $r=2$ (12.5\,Hz, MMAU 59.0), losing 4 MMAU points at $r=4$ (6.25\,Hz). Qwen3-Omni's AuT encoder peaks at $r=2$ (6.25\,Hz, MMAU 61.2). Going to $r=1$ (12.5\,Hz) trades 2 MMAU points for a near-10-point gain on MMAU-Pro instruction-following (67.74 to 77.43) and a 1.8-point gain on the speech subset. AudioSet-Zipformer reaches its best at $r=8$ (6.25\,Hz from 50\,Hz native), four MMAU points above $r=4$.

\providecolor{lightgreen}{HTML}{E8F5E9}

\begin{table*}[t]
\centering
\setlength{\tabcolsep}{\restabcolsep}
\renewcommand{\arraystretch}{\restabrowstretch}
\small

\caption{Effect of projector downsampling rate $r$ across four audio encoders. The audio-token frequency at the LLM input equals the encoder's native frame rate divided by $r$.}
\label{tab:ds_rate}

\resizebox{\restabwidth}{!}{%
\begin{tabular}{@{}l c c c c c c c c c c@{}}
\toprule
\multirow{2}{*}{\textbf{Audio Encoder}} &
\multirow{2}{*}{\textbf{Down.\ $r$}} &
\multicolumn{4}{c}{\textbf{MMAU\;(\textit{test-mini})}} &
\textbf{MMAR} &
\textbf{MMSU} &
\multicolumn{3}{c}{\textbf{MMAU-Pro}} \\
\cmidrule(lr){3-6} \cmidrule(lr){7-7} \cmidrule(lr){8-8} \cmidrule(lr){9-11}
& &
Sound  & Music  & Speech & \textit{Avg.} & \textit{Avg.}
& \textit{Avg.} &
IF & Open-ended & \textit{Avg.}
\\
\midrule

\rowcolor{lightgreen}
\multicolumn{11}{l}{\textit{\textbf{Discriminative pretraining}}} \\
\multirow{2}{*}{AudioSet-Zipformer}     & 4 & \textbf{80.78} & \textbf{69.76} & \textbf{54.05} & \textbf{68.2} & \textbf{54.3} & 47.00 & \textbf{62.91} & 50.84 & \textbf{52.82}  \\
& 2   & 76.58 & 64.37 & 51.35 & 64.1 & 53.0 & \textbf{47.17} & 60.22 & \textbf{55.01} & 49.40\\
\addlinespace[2pt]

\rowcolor{lightgreen}
\multicolumn{11}{l}{\textit{\textbf{ASR-supervised pretraining}}} \\
\multirow{4}{*}{Whisper-large-v2} & \textbf{16} & 67.57	& 59.58	& 49.25 & 58.8 & 48.5 & 43.48 & 58.77 & \textbf{55.67} & \textbf{48.75}\\
& 8 & 69.07	& \textbf{63.17}	& 50.75 & 61.0 & 52.0 & 45.92 & 58.05 & 48.82 & 43.49\\
&4 & \textbf{75.68} & 62.87 & \textbf{60.36} & \textbf{66.3} & \textbf{52.3} & \textbf{50.61} & \textbf{72.57} & 55.39 & 48.40\\
& \textbf{2} & 63.96 & 42.81 & 31.53 & 46.1 & 24.0 & 43.50 & 50.86 & 43.21 & 27.83\\
\addlinespace[2pt]

\rowcolor{lightgreen}
\multicolumn{11}{l}{\textit{\textbf{Joint audio-language pretraining}}} \\
\multirow{2}{*}{Qwen2.5-Omni audio encoder}   & 4 & 68.77 & 50.9 & 45.65 & 55.1 & \textbf{48.3} & 43.31 & \textbf{67.74} & \textbf{53.77} & 45.24 \\
 & 2   & \textbf{70.27} & \textbf{53.29} & \textbf{53.31} & \textbf{59.0} & 48.0 & \textbf{45.41} & 58.05 & 49.46 & \textbf{45.89} \\
\addlinespace[3pt]
\multirow{2}{*}{Qwen3-Omni AuT encoder} & 2 & \textbf{67.87} & \textbf{60.78} & 54.95 & \textbf{61.2} & 48.4 & 45.81 & 67.74 & \textbf{60.59} & 47.32\\
& 1   & 66.97 & 53.89 & \textbf{56.76} & 59.2 & \textbf{49.4} & \textbf{48.28} & \textbf{77.43} & 51.29 & \textbf{50.97}\\
\bottomrule
\end{tabular}%
}
\end{table*}

\section{Scaling Behavior (Cont'd)}
\label{app:scaling-contd}
Table.~\ref{tab:scaling_full} expands Fig.~\ref{fig:exp1} (b) with all subscores. We scale alignment data through three sizes: 576.8K samples (1.6Kh), 1.18M (3.7Kh), and 4.18M (14.2Kh). All samples come from CaptionStew with 10\% speech mixed in. The LLM (Qwen2.5-7B-Instruct) and projector configuration stay fixed.

The headline is saturation on MMAU. Over the full 10$\times$ scale-up, MMAU average drifts by at most 2 points for every encoder (Zipformer 68.20 to 67.30, Whisper 66.30 to 64.30, Qwen2.5-Omni 59.00 to 59.60, Qwen3-Omni 61.20 to 61.40). The Sound and Music subscores show similar saturation; the Speech subset moves more, and we treat it separately below.

Two benchmarks do scale with data. MMAU-Pro Open-ended rises on Zipformer (50.84 to 67.52) and on Qwen2.5-Omni (49.46 to 55.15). MMAR gains across three of the four encoders (Zipformer +3.8, Qwen3-Omni +4.6, Whisper +2.6). Both reward outputs MMAU's closed-set choices cannot absorb (generative answers and event-sequence reasoning), both keep gaining after MMAU has saturated.

A third pattern is mild but consistent: the MMAU Speech subscore degrades with more data on the ASR-supervised and joint encoders (Whisper 60.36 to 52.55, Qwen2.5-Omni 53.31 to 49.75, Qwen3-Omni 54.95 to 52.55), while Zipformer holds. The 10\% speech mixture is fixed in proportion, so absolute speech samples grow with scale. This is dilution, not under-coverage: speech-shaped supervision shrinks relative to an enlarging caption-style corpus.

These patterns match the alignment-ceiling reading of Sec.~\ref{sec:theory-bound}. Data scales the parts of audio understanding that admit richer outputs (MMAU-Pro Open-ended, MMAR). The closed-set ceiling on MMAU is set by the encoder and the LLM, not by alignment data.

\begin{table*}[t]
\centering
\setlength{\tabcolsep}{\restabcolsep}
\renewcommand{\arraystretch}{\restabrowstretch}
\small

\caption{Full scaling results across audio encoders and training-data sizes
(400K, 1M, and 4M from CaptionStew, augmented with 10\% speech). The LLM
(Qwen2.5-7B-Instruct) and projector configuration are held fixed; the projector
downsampling rate $r$ is set per encoder to maintain a post-projector token
rate of 6.25--12.5\,Hz.}
\label{tab:scaling_full}

\resizebox{\restabwidth}{!}{%
\begin{tabular}{@{}l c c c c c c c c c c@{}}
\toprule
\multirow{2}{*}{\textbf{Audio Encoder}} &
\textbf{Training Data} &
\multicolumn{4}{c}{\textbf{MMAU\;(\textit{test-mini)}}} &
\textbf{MMAR} &
\textbf{MMSU} &
\multicolumn{3}{c}{\textbf{MMAU-Pro}} \\
\cmidrule(lr){3-6} \cmidrule(lr){7-7} \cmidrule(lr){8-8} \cmidrule(lr){9-11}
& \# Samples | \# Hours &
Sound  & Music  & Speech & \textit{Avg.} & \textit{Avg.}
& \textit{Avg.} &
IF & Open-ended & \textit{Avg.}
\\
\midrule

\rowcolor{lightgreen}
\multicolumn{11}{l}{\textit{\textbf{Discriminative pretraining}}} \\
\multirow{3}{*}{AudioSet-Zipformer}     & 576.8K | 1.6K & 80.78 & 69.76 & 54.05 & 68.20 & 54.30 & 47.00 & 62.91 & 50.84 & 52.82 \\
& 1.18M | 3.7K   & 80.48 & 67.66 & 53.15 & 67.10 & 57.40 & 54.47 & 67.74 & 58.92 & 51.07 \\
& 4.18M | 14.2K   & 80.28 & 67.37 & 54.25 & 67.30 & 58.10 & 51.81 & 72.57 & 67.52 & 53.19 \\
\addlinespace[2pt]

\rowcolor{lightgreen}
\multicolumn{11}{l}{\textit{\textbf{ASR-supervised pretraining}}} \\
\multirow{3}{*}{Whisper-large-v2}       & 576.8K | 1.6K & 75.68 & 62.87 & 60.36 & 66.30 & 52.30 & 50.61 & 72.57 & 55.39 & 48.40 \\
& 1.18M | 3.7K   & 75.98 & 68.56 & 54.95 & 66.50 & 54.20 & 50.27 & 71.87 & 49.25 & 51.35\\
& 4.18M | 14.2K  & 75.98 & 64.37 & 52.55 & 64.30 & 54.90 & 51.56 & 72.88 & 56.27 & 50.29 \\
\addlinespace[2pt]

\rowcolor{lightgreen}
\multicolumn{11}{l}{\textit{\textbf{Joint audio-language pretraining}}} \\
\multirow{3}{*}{Qwen2.5-Omni audio encoder}   & 576.8K | 1.6K & 70.27& 53.29	& 53.31 & 59.00 & 48.00 & 45.41 & 58.05 & 49.46 & 45.89\\
& 1.18M | 3.7K   & 71.07 & 57.78 & 49.75 & 59.50 & 48.30 & 46.36 & 62.91 & 50.40 & 46.39\\
& 4.18M | 14.2K  & 69.77 & 59.28 & 49.75 & 59.60 & 48.50 & 46.36 & 67.74 & 55.15 & 46.15 \\
\addlinespace[3pt]
\multirow{3}{*}{Qwen3-Omni AuT encoder} & 576.8K | 1.6K & 67.87	& 60.78	& 54.95 & 61.20 & 48.40 & 45.81 & 67.74 & 60.59 & 47.32 \\
& 1.18M | 3.7K   & 67.57 & 60.48 & 54.65 & 60.90 & 50.20 & 44.88 & 67.74 & 59.25 & 47.64\\
& 4.18M | 14.2K  & 69.77 & 61.98 & 52.55 & 61.40 & 53.00 & 42.37 & 62.88 & 62.31 & 48.81\\
\bottomrule
\end{tabular}%
}
\end{table*}

\newpage
\section{Effect of Caption Source (Cont'd)}
\label{app:caption-contd}

In Sec.~\ref{subsubec:caption} of the main text, we analyzed the effect of caption source. Table.~\ref{tab:caption_full} presents a per-subset breakdown, showing each encoder’s performance across different caption sources on the various benchmarks.

\begin{table*}[h]
\centering
\setlength{\tabcolsep}{\restabcolsep}
\renewcommand{\arraystretch}{\restabrowstretch}
\small

\caption{Effect of Caption Source.}
\label{tab:caption_full}

\resizebox{\restabwidth}{!}{%
\begin{tabular}{@{}l c c c c c c c c c c@{}}
\toprule
\multirow{2}{*}{\textbf{Audio Encoder}} &
\multirow{2}{*}{\textbf{Caption Source}} &
\multicolumn{4}{c}{\textbf{MMAU\;(\textit{test-mini})}} &
\textbf{MMAR} &
\textbf{MMSU} &
\multicolumn{3}{c}{\textbf{MMAU-Pro}} \\
\cmidrule(lr){3-6} \cmidrule(lr){7-7} \cmidrule(lr){8-8} \cmidrule(lr){9-11}
& &
Sound  & Music  & Speech & \textit{Avg.} & \textit{Avg.}
& \textit{Avg.} &
IF & Open-ended & \textit{Avg.}
\\
\midrule

\rowcolor{lightgreen}
\multicolumn{11}{l}{\textit{\textbf{Discriminative pretraining}}} \\
\multirow{2}{*}{AudioSet-Zipformer}     & Ground-truth & 80.78 & 69.76 & 54.05 & 68.20 & 54.30 & 47.00 & 62.91 & 50.84 & 52.82\\
                                         & Synthetic   & 79.50& 68.26 & 50.15 & 66.00 & 51.30 & 47.19 & 56.93 & 54.62 & 48.00\\
\addlinespace[2pt]

\rowcolor{lightgreen}
\multicolumn{11}{l}{\textit{\textbf{ASR-supervised pretraining}}} \\
\multirow{2}{*}{Whisper-large-v2}       & Ground-truth & 69.07 &	63.17	& 50.75 & 61.00 & 52.00 & 45.92 & 58.05 & 48.82 & 43.49\\
                                         & Synthetic   & 62.76 &	58.98 &	49.55 & 57.10 & 44.10 & 44.26 & 58.51 & 56.68 & 40.20 \\
\addlinespace[2pt]

\rowcolor{lightgreen}
\multicolumn{11}{l}{\textit{\textbf{Joint audio-language pretraining}}} \\
\multirow{2}{*}{Qwen2.5-Omni audio encoder}   & Ground-truth & 70.27& 53.29	& 53.31 & 59.00 & 48.00 & 45.41 & 58.05 & 49.46 & 45.89\\
                                         & Synthetic  & 50.76& 46.34	& 41.95 & 46.40 & 36.20 & 48.61 & 77.09 & 40.65 & 39.46\\
\addlinespace[3pt]
\multirow{2}{*}{Qwen3-Omni AuT encoder} &  Ground-truth & 67.87	& 60.78	& 54.95 & 61.20 & 48.40 & 45.81 & 67.74 & 60.59 & 47.32 \\
                                         & Synthetic   & 60.96	& 58.68  & 60.96 & 60.20 &  49.80 & 50.43 & 78.72 & 44.22 & 45.09\\
\bottomrule
\end{tabular}%
}
\end{table*}

\section{Samples}
\label{app:sample}

We provide three examples here. The corresponding audio WAV files are available in the \textbf{Supplementary Material}.

\begin{figure*}[ht] 
    \centering
    \label{caption-example}
    \fbox{
        \begin{minipage}{\textwidth}
            
            \textbf{Audio id: WQ-sKXuauTY\_291\_10}\\

             \textbf{Caption:}
             This is a captivating rock ballad featuring a passionate male vocalist. The medium tempo rhythm is anchored by a robust bass line and consistent drumming. The emotional intensity of the song resonates through the powerful use of rock instrumentation, including the bass, drums, and guitar.
            \\
            \\
            \textbf{Response:} 
            That sounds like a compelling rock ballad! To help you further develop or describe this song, here are some elements you might consider:\textbackslash n\textbackslash n1. **Vocal Performance**: The passionate male vocalist is central to the song. Think about the range and emotion he brings to the lyrics. Does he belt out the chorus with intensity, or does he offer softer, more emotive passages in the verses? The vocal delivery can greatly influence the emotional impact of the song.\textbackslash n\textbackslash n2. **Instrumentation**:\textbackslash n   - **Bass Line**: A robust bass line is crucial. Consider the type of bass sound (e.g., clean, distorted) and how it complements the overall mood of the song. Does it provide a solid foundation or add more melodic lines?\textbackslash n   - **Drums**: The consistent drumming provides the heartbeat of the song. Think about the groove and dynamics. Are there moments where the rhythm intensifies or relaxes to build tension and release?\textbackslash n   - **Guitar**: Rock guitars can add a lot of texture and emotion. Do you envision using lead guitar solos to highlight emotional highs? Are there power chords or more intricate riffs to support the song's progression?\textbackslash n\textbackslash n3. **Structure**: A typical rock ballad might follow a verse-chorus-verse-chorus-bridge-chorus structure. How does the song's structure support the emotional arc of the lyrics? Are there instrumental breaks that allow the listener to catch their breath and reflect on the music?\textbackslash n\textbackslash n4. **Lyrics**: The lyrics are a critical component of the song's emotional resonance. What themes or narratives do they convey? Are there specific metaphors or imagery that tie the song together? The lyrics should complement the music and evoke strong emotions.\textbackslash n\textbackslash n5. **Production**: Consider how the production enhances the overall sound. Are there specific effects used on the vocals or instruments? How does the mixing balance the different elements to create a cohesive listening experience?\textbackslash n\textbackslash n6. **Feelings and Themes**: What emotions does the song aim to convey? Love, longing, heartbreak, or resilience? The emotional intensity should be evident in the arrangement and performance.\textbackslash n\textbackslash nBy considering these elements, you can create a rock ballad that not only sounds captivating but also deeply resonates with listeners on an emotional level.
        \end{minipage}
    }
\end{figure*}

\begin{figure*}[ht] 
    \centering
    \label{caption-example}
    \fbox{
        \begin{minipage}{\textwidth}
            
            \textbf{Audio id: nAawjNp4tlk\_12\_10}\\

             \textbf{Caption:}
             A man's voice is heard in the background as a vehicle engine runs. The setting could be various, such as a parking lot or a driveway. No identifiable speech content is discernible from this audio clip.
            \\
            \\
            \textbf{Response:} 
            This audio clip likely depicts the sound of a vehicle engine running in the background with a man's voice heard but no discernible speech content. This could indeed be a common scenario in various settings such as:\textbackslash n\textbackslash n1. **Parking Lot**: A man might be standing or walking around a vehicle while it's running, perhaps checking the oil level or tire pressure.\textbackslash n2. **Driveway**: A man could be getting ready to leave, warming up the vehicle, or perhaps speaking to someone else (not captured in the audio) while the engine is running.\textbackslash n3. **Service Area**: At a mechanic's shop or a car service area, a man might be explaining something to another person while a vehicle's engine is idling.\textbackslash n\textbackslash nWithout more context or additional sounds, it's challenging to pinpoint the exact location, but these are plausible scenarios for the given audio description.

        \end{minipage}
    }
\end{figure*}

\begin{figure*}[ht] 
    \centering
    \label{caption-example}
    \fbox{
        \begin{minipage}{\textwidth}
            
            \textbf{Audio id: \_ULR6ffwADg\_203\_10}\\

             \textbf{Caption:}
             A woman's voice is heard as objects create background noise through clattering and crinkling sounds. The atmosphere remains neutral.
            \\
            \\
            \textbf{Response:} 
            In this scene, the narrative is being conveyed primarily through sound, with the woman's voice standing out against a backdrop of mundane yet evocative sounds. Here’s how the atmosphere and sounds could be described in more detail:\textbackslash n\textbackslash n**Background Sounds:**\textbackslash n- **Clattering:** The sound of dishes or utensils being moved, perhaps in a kitchen. This could be the sound of someone washing dishes, setting the table, or stirring a pot.\textbackslash n- **Crinkling:** This sound suggests paper or thin plastic. It could be the rustle of newspapers being read, the sound of wrapping paper being moved, or the crinkling of bags or packaging.\textbackslash n\textbackslash n**Woman's Voice:**\textbackslash n- The voice should be clear and steady, conveying a sense of routine or possibly a calm demeanor. The tone might indicate someone going about their day or perhaps having a conversation.\textbackslash n\textbackslash n**Atmosphere:**\textbackslash n- Neutral: The overall atmosphere remains unemotional and straightforward, suggesting a typical, everyday setting without any particular tension or excitement.\textbackslash n\textbackslash n**Possible Context:**\textbackslash n- The scene could be set in a home, perhaps a living room or kitchen, where someone is doing household chores or preparing for a task.\textbackslash n- The woman's voice might be speaking about a daily activity or sharing a mundane piece of information.\textbackslash n\textbackslash nThis setup creates a sense of realism and ordinariness, which can be quite effective in storytelling as it often resonates with listeners on a relatable level.

        \end{minipage}
    }
\end{figure*}



\end{document}